\documentclass[pdflatex,sn-mathphys-num,iicol]{sn-jnl}

\usepackage{graphicx}%
\usepackage{placeins}
\usepackage{dblfloatfix}

\usepackage{multirow}%
\usepackage{amsmath,amssymb,amsfonts}%
\usepackage{amsthm}%
\usepackage{mathrsfs}%
\usepackage[title]{appendix}%
\usepackage{xcolor}%
\usepackage{textcomp}%
\usepackage{manyfoot}%
\usepackage{booktabs}%
\usepackage{algorithm}%
\usepackage{algorithmicx}%
\usepackage{algpseudocode}%
\usepackage{listings}%
\usepackage{natbib}
\usepackage{tabularx}
\usepackage{array}
\usepackage{siunitx}

\theoremstyle{thmstyleone}%
\theoremstyle{thmstyletwo}%

\theoremstyle{thmstylethree}%

\begin{document}

\title[What Breaks Scribble-Supervised Segmentation? An Audit-Informed Benchmark for Scribble Annotation Robustness]{What Breaks Scribble-Supervised Segmentation? An Audit-Informed Benchmark for Scribble Annotation Robustness}


\author[1]{\fnm{Minheng} \sur{Wu}}

\author[1]{\fnm{Di} \sur{Wen}}

\author[1]{\fnm{Ruiping} \sur{Liu}}

\author[1]{\fnm{Junwei} \sur{Zheng}}

\author[1]{\fnm{Yufan} \sur{Chen}}

\author[2]{\fnm{Kailun} \sur{Yang}}

\author*[1]{\fnm{Kunyu} \sur{Peng}}
\email{kunyu.peng@kit.edu}

\affil[1]{\orgdiv{Karlsruhe Institute of Technology}, \orgaddress{\street{Kaiserstr. 12}, \city{Karlsruhe}, \postcode{76131}, \state{Baden-Württemberg}, \country{Germany}}}

\affil[2]{\orgdiv{Hunan University}, \orgaddress{\street{No. 2 Lushan South Road}, \city{Changsha}, \postcode{410082}, \state{Hunan}, \country{China}}}
\abstract{Scribble-supervised segmentation is typically evaluated with a fixed annotation set, leaving robustness to distinct semantic and geometric imperfections poorly understood. Motivated by an audit of 12,031 manual PASCAL Visual Object Classes (VOC) 2012 scribbles and prior work, we introduce six controlled perturbation operators. Class corruption, spatial offset, and directional perturbation directly operationalize audit signals. Stroke-length reduction is motivated by prior scribble studies, while random dropout and fragmentation serve as diagnostic controls. Each operator reports requested and achieved changes. We evaluate manual and automatic scribbles across two image domains, two principal backbone settings, and multiple training pipelines, with an additional MiT-B1 PFA anchor evaluation. In the five-seed Tree Energy Loss (TEL)–ResNet101 VOC study, mean intersection over union (mIoU) drops from 70.08 to 1.46 under class corruption, compared with 35.57 for spatial offset, 52.20 for directional perturbation, and 63.24 for stroke-length reduction at their strongest settings. At the evaluated endpoints, semantic-anchor corruption produces the largest degradation, while geometric and support responses depend on the pipeline and dataset. Our benchmark provides a reproducible protocol for diagnosing annotation robustness beyond clean-set accuracy. The operator code is available at \url{https://github.com/Wmhiram99/scribble-operator-benchmark}.}


\maketitle

\section{Introduction}
\label{sec:introduction}

For weakly supervised segmentation, a scribble labels only a few pixels, yet those pixels may supervise an
entire object. This makes scribbles an attractive alternative to costly
dense masks~\citep{papandreou2015weakly,dai2015boxsup,bearman2016point,
lin2016scribblesup}. Modern pipelines propagate sparse labels through regularization~\citep{tang2018regularized}, affinities~\citep{liang2022tree}, pseudo-labels~\citep{pan2021urss}, feature distributions~\citep{wu2023agmm}, or prototypes~\citep{chan2024pfa}. These mechanisms extend sparse anchors beyond labeled pixels and can
propagate the effect of an imperfect anchor.

We summarize scribble imperfections along three complementary properties: semantic identity, spatial geometry, and support/connectivity. An incorrect class changes
\emph{what} an anchor denotes; displacement changes \emph{where}
supervision enters; direction alters local geometry; and shortening,
dropout, or fragmentation modify support and connectivity.  We instantiate these properties through controlled perturbation operators, which transform clean scribbles into matched noisy training labels and together define the proposed robustness benchmark.
Existing work largely optimizes a fixed scribble set or studies selected variations in class-aware cues~\citep{zhang2024scribblehides}, generation~\citep{schnell2023scribblegen}, and drawing style~\citep{boettcher2024scribblesforall,zhang2025cspnet}. General noisy-label pipelines focus mainly on class corruption, while segmentation robustness benchmarks typically perturb the input image rather than the training annotation~\citep{patrini2017losscorrection,liu2022adele,kamann2020robustseg}. Perturbing the training scribbles is necessary here because they define the sparse supervisory anchors whose semantic, geometric, and support errors are propagated during learning.
A unified, audit-grounded benchmark for decomposing scribble imperfections is therefore still missing. To ground such a benchmark empirically, we first audit discrepancies between manual scribbles and dense references, then translate the observed modes into controlled perturbations with measurable achieved values.

Our research questions are:
\textbf{RQ1:} How can observed discrepancies between scribble annotations and dense references be translated into an audit-grounded benchmark of controlled scribble perturbations?
\textbf{RQ2:} Which semantic, geometric, and support properties of scribble annotation most strongly affect segmentation?
\textbf{RQ3:} Which response patterns persist across scribble annotation sources, datasets, backbones, and learning pipelines?

\begin{figure*}[!t]
    \centering
    \includegraphics[
        width=0.80\textwidth
    ]{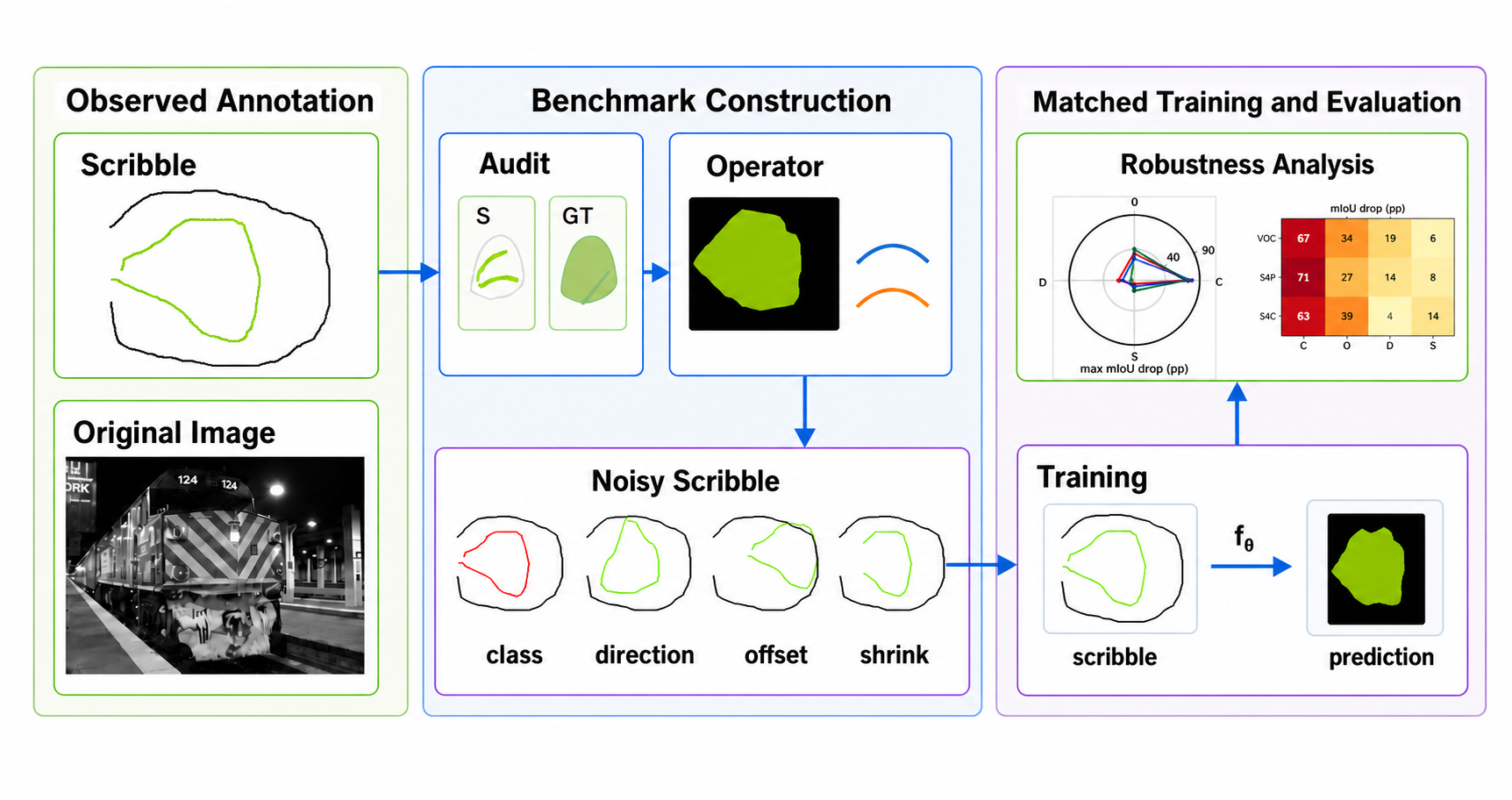}
    \caption{Noisy scribble semantic segmentation benchmark. Audit signals motivate controlled annotation perturbations, whose achieved properties are verified before matched training and robustness evaluation. The inset reports maximum mIoU drops in percentage points for TEL--ResNet101 across the nine ratios. VOC2012 uses the single-run full-ratio path; S4Pascal and S4Cityscapes are descriptive single-run paths. C, O, D, and S denote Class, Offset, Direction, and Shrink}
    \label{fig:overview}
\end{figure*}

To investigate these research questions, we introduce an \emph{audit-to-stress-test benchmark}, whose overall workflow is illustrated in Figure~\ref{fig:overview}. We audit $12{,}031$ manual PASCAL VOC2012 scribbles~\citep{everingham2010pascal}. The audit directly motivates class corruption, spatial offset, and directional perturbation. Stroke-length reduction is informed by prior scribble research, while random dropout and fragmentation are included as diagnostic controls. These six operations are implemented as controlled perturbation operators. The first four form the cross-setting benchmark because they have direct audit evidence or provide relatively isolated control of annotation properties.

For each operator, ratios from $1.0$ (the clean identity setting) to $0.1$ request progressively stronger perturbations to the scribble labels. The achieved change can differ from the request because label maps are discrete and each operator has its own score definition. These labels are used to train the evaluated segmentation pipelines, whose predictions on validation RGB images are measured against dense ground truth. We analyze the resulting mIoU responses as operator-specific robustness profiles, extend the evaluation across datasets, backbones, and learning pipelines, and repeat selected settings with multiple seeds to assess the stability of the primary findings.
Our contributions can be summarized as:
\begin{enumerate}
    \setlength{\itemsep}{1pt}
    \setlength{\topsep}{2pt}

    \item A dataset-scale audit of $12{,}031$ PASCAL VOC2012 scribbles that directly motivates three perturbation operators and informs a broader six-operation benchmark with achieved-value diagnostics.
    \item A controlled analysis of how semantic, geometric, and support properties affect segmentation robustness under operator-specific perturbations.
    \item Repeated-seed and cross-setting checks showing which response patterns are retained across annotation sources, datasets, backbones, and learning pipelines.
\end{enumerate}

\section{Related Work}

\paragraph{Scribble-supervised segmentation.}
ScribbleSup established scribbles as an efficient weak annotation by alternating graphical label propagation with network learning~\citep{lin2016scribblesup}.
Subsequent methods propagate sparse evidence through random-walk propagation~\citep{vernaza2018randomwalk}, graph- and energy-based regularization~\citep{tang2018regularized,tang2018ncut}, boundary guidance and progressive inference~\citep{wang2019bpg,xu2021psi}, and uncertainty-aware self-training~\citep{pan2021urss}.
Recent methods further use Transformer representations and class-conditional modeling~\citep{su2023sasformer,wu2023agmm}, tree-structured affinities~\citep{liang2022tree}, or prototype-based feature augmentation~\citep{chan2024pfa}.
Despite their different mechanisms, these methods share a structural dependency. Labeled scribble pixels act as semantic and spatial anchors from which supervision is propagated.
Existing scribble-supervised methods mainly improve learning under a given annotation set through propagation, regularization, or representation design~\citep{lin2016scribblesup,pan2021urss,liang2022tree,chan2024pfa}. Rather than improving learning for a fixed scribble set, we introduce an audit-informed benchmark that measures how controlled perturbations of the sparse label change segmentation performance, yielding operator-specific robustness profiles.

\paragraph{Scribble sources and annotation variation.}
Manual ScribbleSup~\cite{lin2016scribblesup} annotations preserve annotator-dependent variation in placement, length, and shape, but are largely confined to PASCAL VOC.
Scribbles for All~\cite{boettcher2024scribblesforall} generates reproducible scribbles from dense masks across multiple 2D datasets, whereas ScribbleKITTI~\cite{unal2022scribblekitti} extends sparse supervision to 3D LiDAR scenes. 
Other studies examine class-aware cues~\citep{zhang2024scribblehides}, automatic scribble generation and augmentation~\citep{schnell2023scribblegen}, and variation in drawing style~\citep{zhang2025cspnet}. 
Interactive segmentation also studies tolerance to erroneous scribbles or clicks, but under online correction rather than offline scribble-supervised training~\citep{bai2014errortolerant,zhang2025ntclick}. 
These efforts broaden annotation sources and styles but lack an audit-grounded protocol for separately measuring semantic, spatial, directional, support, and connectivity sensitivity within a common scribble source. Our audit-informed benchmark separates semantic, geometric, support, and connectivity sensitivity rather than treating scribble quality as a single variable.

\paragraph{Noisy annotations and robustness evaluation}
Noisy-label learning largely targets incorrect class assignments through noise-transition modeling and loss correction~\citep{sukhbaatar2015noisy,patrini2017losscorrection}, robust learning objectives~\citep{zhang2018gce}, and sample-selection or collaborative-training strategies~\citep{han2018coteaching,arazo2019unsupervised,northcutt2021confident}.
A broader overview of these directions is provided by Song et al.~\citep{song2020noisylabelsurvey}.
Segmentation additionally introduces spatially structured errors affecting pixels, regions, and boundaries~\citep{liu2022adele,rottmann2023labelerrors,li2023diverse}.
Scribble noise is distinct because sparse labels are subsequently propagated. The same number of altered pixels can have different effects depending on what they label, where they lie, and how much coherent support remains. Corruption benchmarks systematically vary image perturbations while holding annotations fixed~\citep{kamann2020robustseg}. We transpose this principle to the training annotation. Our benchmark audits $12,031$ manual VOC scribbles and uses the supported disagreement patterns to motivate class, offset, and direction operators. Stroke-length reduction is informed by prior work, while dropout and fragmentation serve as diagnostic controls. All operations record requested and achieved properties. This design enables reproducible robustness analysis across annotation sources, datasets, backbones, and learning pipelines.

\section{Benchmark Methodology}
\label{sec:methodology}
\raggedbottom

\subsection{Problem setup}

Let $I_i$ denote an image, $S_i$ its sparse scribble annotation, and $Y_i$ the corresponding dense reference. Here, clean refers to the original, unperturbed scribble map and does not imply perfect agreement with the dense reference.
We measure how a segmentation pipeline responds when one property of $S_i$ is altered while the image is fixed.
Dense references are only used to audit the original annotations and to construct and measure controlled scribbles. They are never used as training targets. Although noisy scribble maps are also generated for validation images, validation does not consume them. The model predicts from validation RGB images, and fixed dense ground-truth (GT) masks are used to compute validation loss, mIoU, and checkpoint selection.

\subsection{Corpus audit}

We start from the manual PASCAL VOC scribbles because they retain variation introduced during annotation.
The audit covers $12{,}031$ images: $10{,}582$ from the augmented training set and $1{,}449$ from the VOC validation set.
The scribbles and the VOC/Semantic Boundaries Dataset (SBD) references share the same image coordinates and label taxonomy~\citep{everingham2010pascal,hariharan2011semantic}.
Automated screening compares them at pixel and connected-component levels.
The signals capture class disagreement, foreground--background leakage, spatial disagreement, a scribble class that is absent from the dense reference, and axial direction discrepancy.
The signals overlap, so the counts reported in Table~\ref{tab:audit} are not additive.

\begin{table}[t]
\centering
\footnotesize
\setlength{\tabcolsep}{5pt}
\renewcommand{\arraystretch}{1.12}

\caption{Screening results for manual PASCAL VOC2012 scribbles. The first five rows are image-level signals and may co-occur in one image. The direction analysis uses $17{,}937$ eligible foreground components of $28{,}227$ total.}
\label{tab:audit}
\begin{tabularx}{\columnwidth}{@{}Xrr@{}}
\toprule
\textbf{Screening signal} & \textbf{Count} & \textbf{Share (\%)} \\
\midrule
\multicolumn{3}{@{}l}{\textit{Image-level signals} ($n=12{,}031$)} \\
Foreground-class disagreement        & 1,556 & 12.93 \\
Foreground scribble on GT background  & 7,272 & 60.44 \\
Background scribble on GT foreground  & 1,328 & 11.04 \\
Spatial-disagreement candidate        & 1,831 & 15.22 \\
Scribble class absent from GT         &    15 &  0.12 \\
\addlinespace[0.4em]
\multicolumn{3}{@{}l}{\textit{Eligible foreground components} ($n=17{,}937$)} \\
Direction discrepancy ($\geq45^\circ$) & 1,437 &  8.01 \\
\bottomrule
\end{tabularx}

\end{table}

For each image-level signal, we inspected up to 50 flagged images. When fewer than 50 were flagged, all available images were checked. For the component-level direction signal, we inspected 50 flagged components. All 15 images flagged for ``Scribble class absent from GT'' were checked. This was a qualitative sanity check of the screening predicates, not an estimate of screening accuracy.
The gap between image-level and pixel-level frequency matters most for the leakage signal: $7{,}272$ images contain at least one foreground scribble pixel on GT background, but such pixels are only $2.955\%$ of the eligible scribble pixels.
The audit identifies properties worth testing, not the frequency of annotation errors.

\subsection{Intervention axes}

The audit motivates semantic identity, spatial alignment, and trajectory direction. Prior work motivates coherent support~\citep{boettcher2024scribblesforall}. These four properties define the primary cross-setting benchmark.
Dropout and Fragmentation remain diagnostic controls because they couple support and connectivity rather than isolating one property. We distinguish isolated control, which preserves other measured properties where the discrete label map permits, from coupled control, whose construction necessarily changes an additional property. Class is isolated and Shrink approximately isolated. Offset, Direction, Dropout, and Fragmentation are coupled through translation, rotation, distributed removal, or cutting. Table~\ref{tab:operator-scope} summarizes the six operators.

\begin{table}[!t]
\centering
\footnotesize
\setlength{\tabcolsep}{10pt} 
\renewcommand{\arraystretch}{1.12}

\caption{The six perturbation operators and their target properties. The primary set spans four annotation properties. Dropout and Fragmentation are diagnostic controls. We distinguish isolated from coupled operational control.}
\label{tab:operator-scope}
\begin{tabular*}{\columnwidth}{@{\extracolsep{\fill}}lll@{}}
\toprule
\textbf{Operator} & \textbf{Basis} & \textbf{Target property} \\
\midrule
\multicolumn{3}{@{}l}{\textit{Primary operators}} \\
Class      & Corpus audit & Semantic identity \\
Offset     & Corpus audit & Spatial alignment \\
Direction  & Corpus audit & Trajectory direction \\
Shrink     & Prior work   & Coherent support \\
\addlinespace[0.4em]
\multicolumn{3}{@{}l}{\textit{Diagnostic operators}} \\
Dropout       & Prior work & Distributed support \\
Fragmentation & Designed   & Connectivity \\
\bottomrule
\end{tabular*}

\end{table}

\subsection{Operator-specific control}

The six interventions act on different properties, so a shared corruption budget would be misleading.
We instead give each operator $t$ its own measurement $Q_t$.
For a clean scribble $S_i$ and a requested ratio $r\in\{0.9,0.8,\ldots,0.1\}$, the operator produces $S_{i,t,r}=T_t(S_i,Y_i;r)$, where $T_t$ is the construction rule for operator $t$.  Each diagnostic $Q_t(S;S_i,Y_i)$ takes a candidate scribble map $S$ as its first argument, with $S=S_i$ for the clean condition and $S=S_{i,t,r}$ otherwise.
We record
{\setlength{\abovedisplayskip}{3pt}
\setlength{\belowdisplayskip}{3pt}
\setlength{\abovedisplayshortskip}{3pt}
\setlength{\belowdisplayshortskip}{3pt}
\begin{equation}
\begin{aligned}
q^0_{i,t}      &= Q_t(S_i;S_i,Y_i), \\
\tau_{i,t,r}   &= G_t(r;q^0_{i,t}), \\
\hat q_{i,t,r} &= Q_t(S_{i,t,r};S_i,Y_i), \\
e_{i,t,r}      &= \lvert \hat q_{i,t,r}-\tau_{i,t,r}\rvert .
\end{aligned}
\label{eq:requested-achieved}
\end{equation}
}
Here $q^0_{i,t}$ is the clean measurement, $G_t$ is the operator-specific mapping from nominal ratio to requested value, $\tau_{i,t,r}$ is the requested value, $\hat q_{i,t,r}$ is the achieved value, and $e_{i,t,r}$ is their discrepancy.
The clean condition is indexed by $r=1.0$.
Within an operator, a lower ratio requests a stronger intervention.
The same ratio has no common meaning across operators, so cross-operator comparisons describe model response, not matched severity.
Unless required by dataset-specific label encoding, the operator constants are fixed across datasets, backbones, and pipelines.  Each operator--ratio condition is generated independently from the same clean scribble map, rather than by composing perturbations sequentially.

\subsection{Perturbation operators}

\paragraph{Class corruption.}
We divide $S_i$ into $K_i$ valid 8-connected components $R_k$.  The component agreement and class diagnostic are
\begin{equation}
\begin{aligned}
a_k(S) &= \frac{\lvert\{p\in R_k:S(p)=Y_i(p)\}\rvert}{\lvert R_k\rvert},\\
Q_{\mathrm{class}}(S;S_i,Y_i)
&=\frac{1}{\sum_{k=1}^{K_i}\lvert R_k\rvert}
\sum_{k=1}^{K_i}\lvert R_k\rvert\,a_k(S)\\
&\qquad\times\mathbf{1}\!\left[a_k(S)\geq0.8\right].
\end{aligned}
\label{eq:component-agreement}
\end{equation}
Here $p$ indexes pixels and $\mathbf{1}[\cdot]$ is the indicator function. Components satisfying the eligibility criterion are selected to approach the requested target. The selected components are relabeled within the valid class set for the dataset, while their coordinates and support remain fixed. The achieved value is recomputed after relabeling. Since components are indivisible, the requested value may not be attained exactly.

\paragraph{Spatial offset.}
Offset changes where a scribble anchors supervision.
For each foreground GT component $m$, let $c^{Y}_{m}$ be its centroid and $\rho_m$ its maximum distance from the centroid to a component pixel. Let $c^{S}_{m}$ be the centroid of the scribble components that overlap this GT component, weighted by component area and a soft overlap weight. The GT-component weights are $w_m=|Y_m|/\sum_j|Y_j|$, so they sum to one. For the background term, $c^{Y}_{\mathrm{bg}}$ is the centroid of the GT background, $c^{S}_{\mathrm{bg}}$ is the corresponding overlap-weighted scribble centroid, and $\rho_{\mathrm{bg}}=\operatorname{clip}(1.4\,\bar\rho,20,90)$ pixels, where $\bar\rho$ is the mean foreground-component radius.
A bounded radial kernel maps a normalized displacement to a similarity,
\begin{equation}
\phi(x)=\max\!\left(0.02,\,[1-\operatorname{clip}(x,0,1)]^{2.2}\right),
\label{eq:offset-kernel}
\end{equation}
and the offset score combines matched-pair and background terms,
\begin{equation}
\begin{aligned}
Q_{\mathrm{offset}}
  &= (1-\beta)\sum_{m} w_m\\
  &\quad\times\phi\!\left(\frac{\lVert c^{S}_{m}-c^{Y}_{m}\rVert}{1.25\,\rho_m}\right)\\
  &\quad + \beta\,\phi\!\left(\frac{d_{\mathrm{bg}}}{1.25\,\rho_{\mathrm{bg}}}\right),
\end{aligned}
\label{eq:offset-score}
\end{equation}
where $d_{\mathrm{bg}}=\lVert c^{S}_{\mathrm{bg}}-c^{Y}_{\mathrm{bg}}\rVert$ and $\beta$ is the background fraction of labeled scribble pixels, clipped to $[0.10,0.25]$. The foreground-centroid estimate uses soft overlap weights that increase smoothly from zero to one as a scribble component's overlap with a GT component increases from $0.05$ to $0.30$.
Boundary-safe translation candidates preserve labels, avoid canvas clipping, and move the score toward its target.  Their overlap, leakage, and boundary-contact changes are retained.

\paragraph{Directional perturbation.}
Direction compares the principal axes of foreground scribble and GT components of the same class.  Each scribble component is assigned to the same-class GT component with maximum overlap; unmatched components are omitted.  The $45^\circ$ threshold is used by the audit screen, not as a generator-eligibility rule.  Principal component analysis (PCA) yields unit principal axes $u_{S,g}(\alpha_g)$ and $u_{Y,g}$ for the rotated scribble and matched GT component; opposite signs represent the same orientation.  For the $G_i$ retained matched groups and their selected rotations $\alpha_g$,
\begin{equation}
\begin{aligned}
\Delta\theta_g(\alpha_g)
&= \arccos\!\left(\lvert u_{S,g}(\alpha_g)^{\top}
u_{Y,g}\rvert\right)\\
&\in[0,90^\circ],\\
Q_{\mathrm{dir}}(S;S_i,Y_i)
&= \frac{1}{\sum_{g=1}^{G_i}w_g}
\sum_{g=1}^{G_i} w_g\,\gamma_g(\alpha_g)\\
&\qquad\times\exp\!\left[-\omega_g\!\left(
\frac{\Delta\theta_g(\alpha_g)}{25^\circ}\right)^2\right],
\end{aligned}
\label{eq:axial-angle}
\end{equation}
where $w_g$ is the group weight, $\gamma_g(\alpha_g)$ is its feasible-angle geometry factor, and $\omega_g$ is its GT-anisotropy gate.  $\omega_g$ depends only on the matched GT component and is constant across candidate rotations.
Eligible groups require scribble/GT areas of at least $15/30$ pixels and cover at least $0.5\%$ of valid scribble support; at most the $12$ largest are optimized. PCA uses skeleton support and falls back to raster support when fewer than five skeleton points are available. The GT anisotropy gate is zero below $1.25$ rather than excluding the group. Candidate rotations span $[-90^\circ,90^\circ]$ in $6^\circ$ steps, and two coordinate-descent passes approach the requested score.
Rotation preserves class identity but can alter overlap or leave the matched region.

\paragraph{Stroke-length reduction.}
Shrink removes coherent support while preserving stroke character.  Each eligible component is skeletonized, trimmed from approximate diameter endpoints according to its area allocation, and reconstructed at approximately its original width.  If $R'_k$ is the retained part of component $R_k$, the achieved support is
\begin{equation}
Q_{\mathrm{shrink}}=\frac{\sum_k \lvert R'_k\rvert}{\sum_k \lvert R_k\rvert}.
\label{eq:shrink}
\end{equation}
The remaining pixels retain their class labels.  Shrink is an approximately isolated coherent-support intervention, unlike Dropout, whose distributed sampling can also alter connectivity.

\paragraph{Random dropout.}
Random dropout reduces distributed support through seeded selection over skeleton support, rather than by independently deleting original raster pixels.
The implementation takes a seed argument (default $0$), skeletonizes the eligible scribble support, shuffles skeleton points, and reconstructs the selected points at approximately the original stroke width. With $A(S_i)$ denoting the eligible scribble pixels and $A(S_{i,t,r})$ the surviving set,
\begin{equation}
Q_{\mathrm{drop}}=\frac{\lvert A(S_{i,t,r})\rvert}{\lvert A(S_i)\rvert}.
\label{eq:dropout}
\end{equation}
Surviving support keeps its original labels and local coordinates.  Reconstruction can split or merge raster connectivity, so Dropout is a coupled distributed-support diagnostic.

\paragraph{Fragmentation.}
Each scribble component is assigned to the GT region with maximal overlap, and $k_j(S)$ is the number assigned to region $j$.  Let $\mathcal{F}_i$ denote covered foreground regions and $\mathcal{B}_i$ background regions; within each set, $w_j$ is proportional to the GT-region area.  The implemented fluency score is
\begin{equation}
\begin{aligned}
q_{\mathrm{fg}} &= \sum_{j\in\mathcal{F}_i}w_j\exp[-0.4(k_j(S)-1)],
\\
q_{\mathrm{bg}} &= \sum_{j\in\mathcal{B}_i}w_j\exp[-0.8k_j(S)],\\
Q_{\mathrm{frag}}(S;S_i,Y_i) &= 0.7q_{\mathrm{fg}}+0.3q_{\mathrm{bg}}.
\end{aligned}
\label{eq:fragmentation-score}
\end{equation}
Here, $q_{\mathrm{fg}}$ penalizes multiple scribble components assigned to a covered foreground GT region and uncovered foreground regions are excluded. The term $q_{\mathrm{bg}}$ penalizes scribble components assigned to background GT regions, with no assigned component yielding a local factor of one. The region weights are proportional to GT area. Candidate cuts are selected to approach this target.  The operation therefore tests whether separating local stroke pieces changes learning beyond support reduction alone.  Because every cut can also remove support, Fragmentation is a coupled diagnostic control rather than a pure connectivity intervention.

\begin{figure*}[t]
\centering
\includegraphics[width=0.90\textwidth]{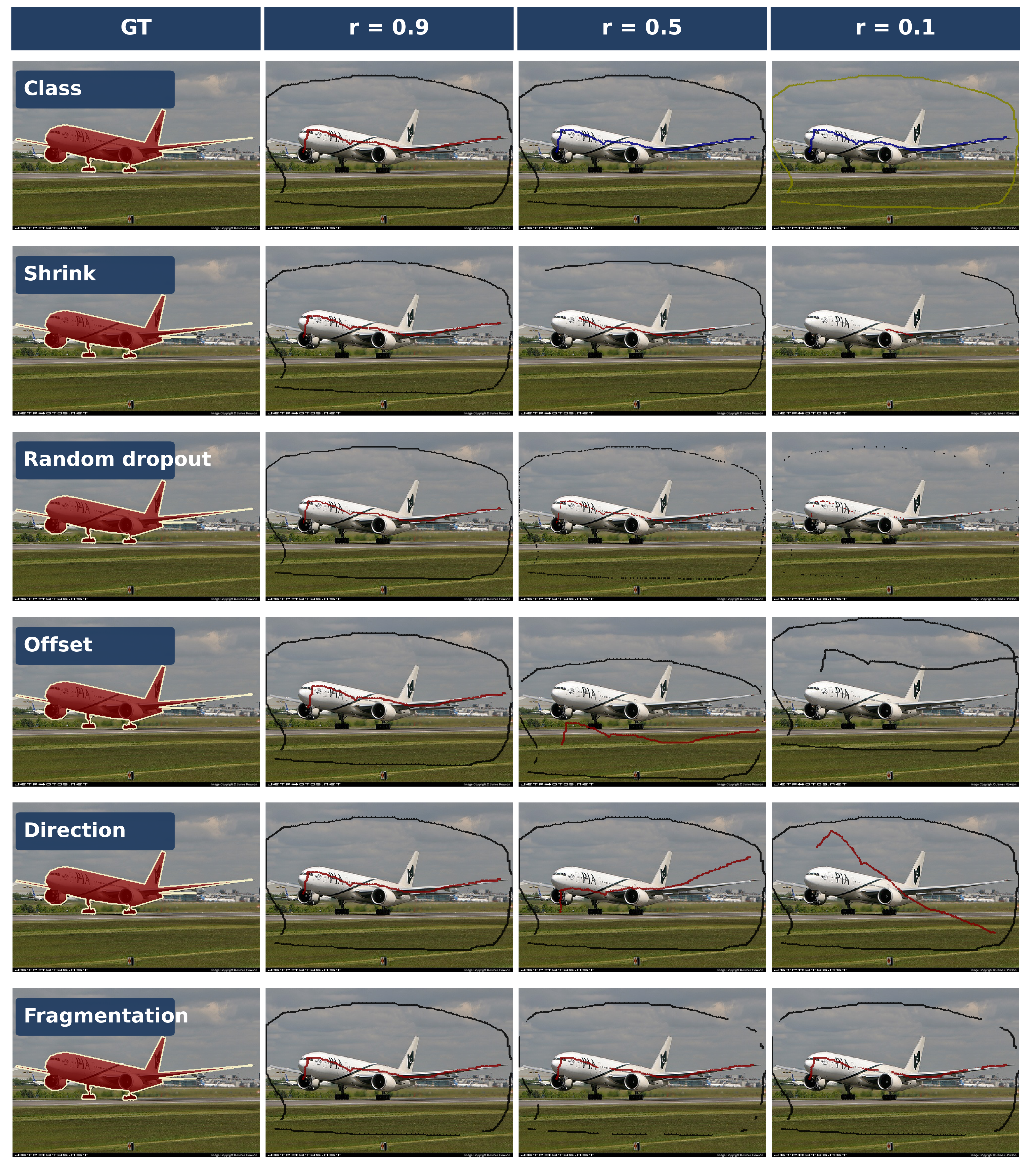}
\caption{Six manual-VOC operations on image 2007\_000256~\citep{everingham2010pascal}. The leftmost column shows the GT for visual reference; the remaining columns show generated labels at $r=0.9,0.5,0.1$. In the class row, relabeling denotes replacement of a selected whole component's class. Per-operator generation success and achieved-value distributions are reported in the supplementary material}
\label{fig:operators}
\end{figure*}

Figure~\ref{fig:operators} shows the six operators on a common image at three requested ratios, with lower $r$ indicating a stronger intervention within a row.

Shrink, Dropout, and Fragmentation all reduce support, but respectively retain, distribute, or split it. Their comparison tests whether response depends on support amount or its organization.

\begin{figure*}[!b]
\centering
\includegraphics[width=0.82\textwidth]{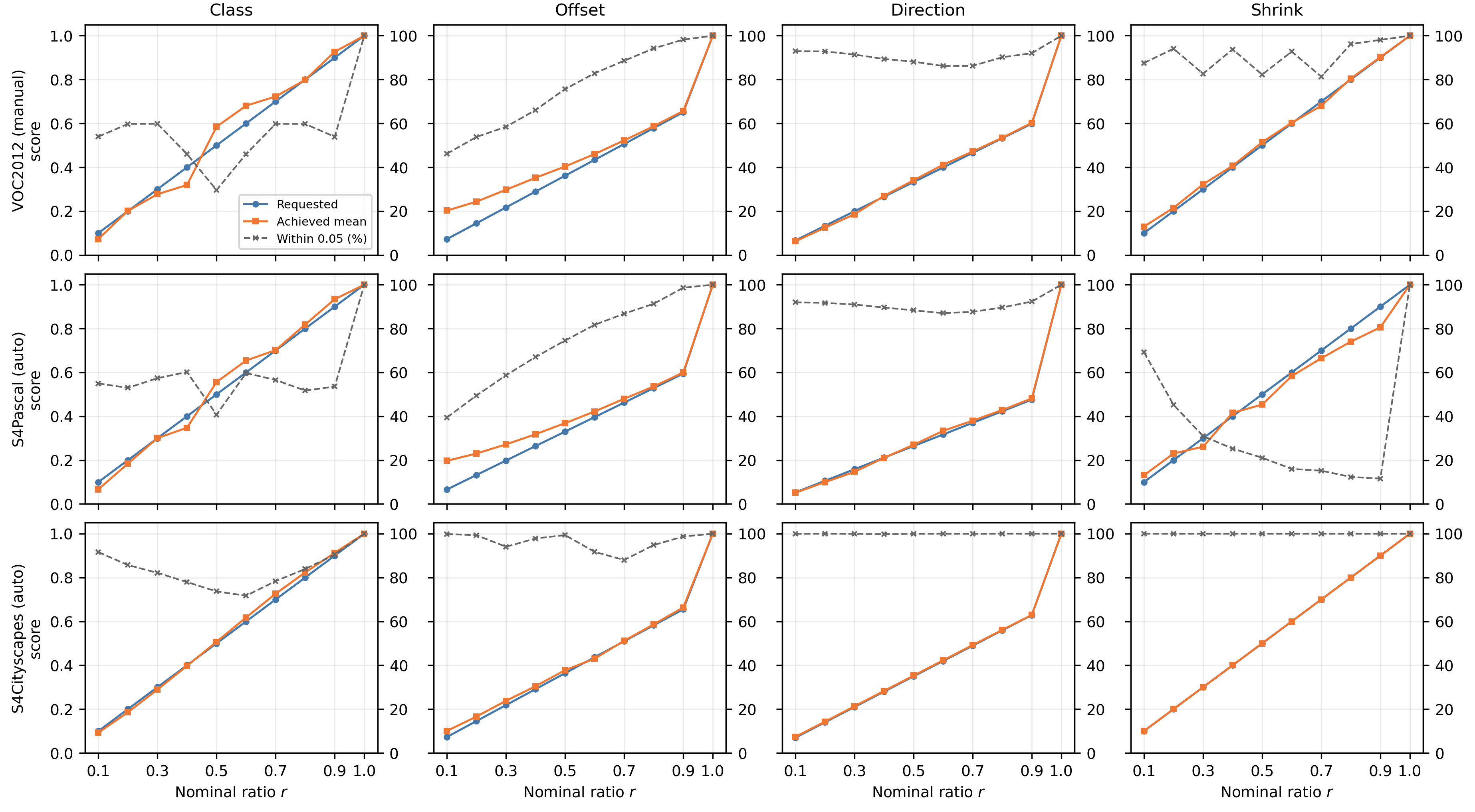}
\caption{Requested and achieved operator scores and image-level target attainment. Each panel is one dataset--operator pair. The left axis shows the requested score and achieved mean; the right axis shows the percentage of images with absolute score error at most $0.05$. The clean/identity reference at $r=1.0$ is shown for orientation. Its perfect attainment is expected by construction and it is excluded from perturbed-level summaries. Scores remain operator-specific}
\label{fig:operator-reachability}
\end{figure*}

\subsection{Operator reachability and target attainment}
\label{subsec:operator-reachability}

A nominal ratio is a request to an operator, not a guarantee that every discrete label map can attain the requested score. For each dataset--operator--ratio condition, we report the source records' mean requested score $\tau_{t,r}$, mean achieved score $\hat q_{t,r}$, and image-level target-attainment rate
\begin{equation}
 \begin{aligned}
 A_{t,r}(\epsilon)&=\frac{1}{N_{t,r}}\sum_{i=1}^{N_{t,r}}
 \mathbf{1}\!\left[e_{i,t,r}\leq\epsilon\right],\\
 \epsilon&=0.05.
 \end{aligned}
 \label{eq:target-attainment}
\end{equation}

Figure~\ref{fig:operator-reachability} plots each condition's requested score, achieved mean, and attainment rate without interpolation.

Here $e_{i,t,r}=|\hat q_{i,t,r}-\tau_{i,t,r}|$ is the absolute score error for image $i$, and $N_{t,r}$ counts images with valid paired requested and achieved diagnostics. The attainment rate measures score calibration, not the fraction of generator jobs that completed successfully. The requested and achieved scores retain each generator's original definition. Offset is a GT-guided composite score and Direction is a trajectory-consistency score, not a displacement in pixels or a rotation angle. Shrink measures retained support. The plotted $r=1.0$ row is an identity reference, not a generated perturbation. Its requested and achieved values are set to 1.0 for display, so its perfect attainment is by construction; it is shown for orientation and excluded from every perturbed-level summary. Since score scales differ, calibration is compared only within an operator and dataset.

The diagnostic records cover all $12{,}031$ images in each of manual VOC and S4Pascal, and all $2{,}975$ S4Cityscapes images. Across the nine perturbed levels, Direction has comparatively high average attainment (89.9\% for manual VOC, 89.9\% for S4Pascal, and 99.9\% for S4Cityscapes), whereas Class and Offset vary more by dataset, with Class ranging from 52.1\% to 81.7\% and Offset from 72.0\% to 96.0\%. Shrink is particularly dataset-sensitive, with mean attainment of 89.9\% for manual VOC, 27.4\% for S4Pascal, and 100\% for S4Cityscapes. S4Pascal has substantially lower Shrink attainment than S4Cityscapes. Thus a common nominal ratio does not imply a common achieved change, even for the same operator across annotation sources.

Table~\ref{tab:operator-reachability} shows that target attainment depends on both the operation and the annotation source. Direction attains its requested score consistently across the three datasets, while Class and Shrink show larger dataset-dependent differences. In particular, Class attainment is lower for manual VOC (52.1\%) and S4Pascal (54.2\%) than for S4Cityscapes (81.7\%), and Shrink attainment ranges from 27.4\% for S4Pascal to 100\% for S4Cityscapes. Since each operator retains its own score definition, these attainment rates are interpreted within each operator.

\begin{table*}[!t]
\centering\small
\setlength{\tabcolsep}{4pt}

\caption{Generator target reachability over the nine perturbed levels $r=0.1$--$0.9$. Mean level absolute error is the mean across the nine conditions of the absolute difference between achieved and requested mean scores; mean target attainment is the mean image-level fraction with absolute score error no greater than 0.05. The $r=1.0$ identity/clean reference is excluded. Scores are operator-specific and are not physical translations or angles.}
\label{tab:operator-reachability}
\resizebox{\textwidth}{!}{%
\begin{tabular}{llrrrrr}
\toprule
Dataset & Operator & Images & Mean level abs. error & Mean within-0.05 (\%) & $r=.1$ achieved & $r=.1$ within-0.05 (\%) \\
\midrule
VOC2012 (manual) & Class & 12,031 & 0.039 & 52.1 & 0.073 & 54.0 \\
VOC2012 (manual) & Offset & 12,031 & 0.052 & 73.8 & 0.203 & 46.2 \\
VOC2012 (manual) & Direction & 12,031 & 0.007 & 89.9 & 0.062 & 92.9 \\
VOC2012 (manual) & Shrink & 12,031 & 0.013 & 89.9 & 0.130 & 87.6 \\
\addlinespace[2pt]
S4Pascal (auto) & Class & 12,031 & 0.030 & 54.2 & 0.067 & 55.0 \\
S4Pascal (auto) & Offset & 12,031 & 0.050 & 72.0 & 0.197 & 39.4 \\
S4Pascal (auto) & Direction & 12,031 & 0.007 & 89.9 & 0.051 & 92.0 \\
S4Pascal (auto) & Shrink & 12,031 & 0.041 & 27.4 & 0.131 & 69.2 \\
\addlinespace[2pt]
S4Cityscapes (auto) & Class & 2,975 & 0.014 & 81.7 & 0.093 & 91.6 \\
S4Cityscapes (auto) & Offset & 2,975 & 0.013 & 96.0 & 0.100 & 99.8 \\
S4Cityscapes (auto) & Direction & 2,975 & 0.003 & 99.9 & 0.074 & 99.9 \\
S4Cityscapes (auto) & Shrink & 2,975 & 0.000 & 100.0 & 0.100 & 100.0 \\
\bottomrule
\end{tabular}
}
\end{table*}

The lower Class attainment in the two Pascal settings is consistent with the discrete structure of this intervention. Class recolors an entire scribble component, so the achieved score changes in component-sized steps rather than continuously. We counted 8-connected components separately within each non-background semantic label in the raster annotations. The mean count per image was 2.35 (median 2) for manual VOC and 4.09 (median 3) for S4Pascal, so many images have only a few independently relabelable units. These counts provide a reproducible raster proxy for scribble units and do not imply that every component is exactly one pen stroke. With few units, attainable scores are coarse and a requested ratio can fall between attainable values. This offers a plausible explanation for missed Class targets when the smallest possible change already recolors one whole component. It is consistent with, but does not prove, the mechanism behind the dataset difference.

Accordingly, downstream comparisons should consider achieved properties as well as nominal ratios. Table~\ref{tab:operator-reachability} reports the condition-level mean absolute difference, computed as the mean of the nine perturbed ratio-level mean-score errors, and the mean image-level attainment. The latter is computed within each ratio from valid paired records and then averaged over the nine perturbed levels. Together, these summaries show aggregate calibration error and the fraction of images meeting the tolerance. The $0.05$ tolerance is an operational calibration criterion, not a test of successful job completion.

\subsection{Cross-property changes within a target-operator output}
\label{subsec:operator-confounding}

\begin{figure*}[!t]
\centering
\includegraphics[width=0.88\textwidth]{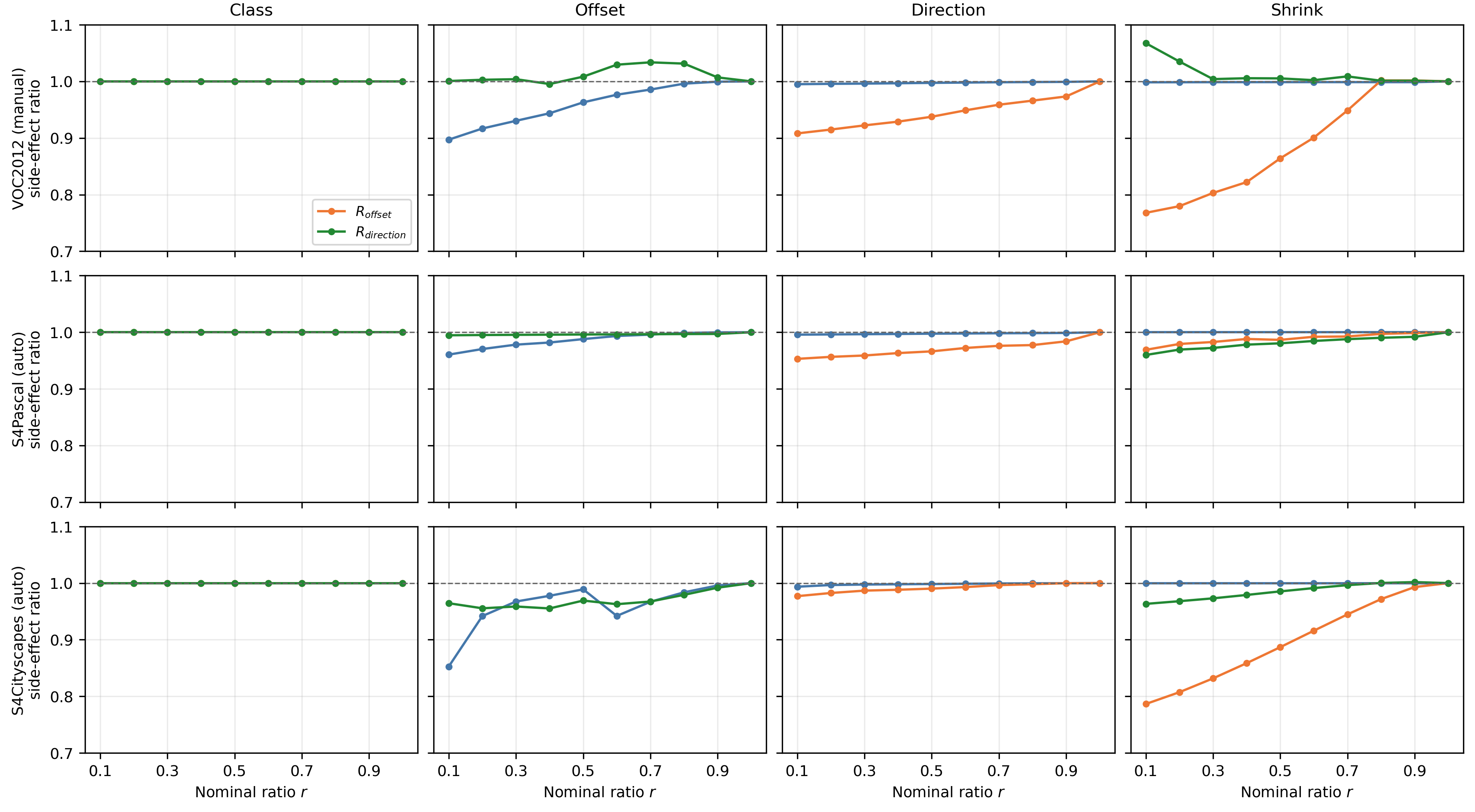}
\caption{Cross-property score ratios in each target-operator output over $r=0.1$--$0.9$. Each column identifies the target operation and each row a dataset. Only non-target properties are plotted; the dashed horizontal line marks ratio one. The vertical axis is restricted to the observed near-neutral range to magnify small deviations. Distances from the reference should therefore be interpreted using the numeric values, not as a full-scale visual effect size. Score ratios use the report-specific fixed-binding or native score definitions described in the text}
\label{fig:operator-side-effects}
\end{figure*}

To measure operational coupling, we compare each output $S_{i,t,r}$ only with its own input scribble $S_i$. We do not substitute an independently generated output from another operator. Let $U_i=\{p:S_i(p)\neq255\}$ and $V_{i,t,r}=\{p:S_{i,t,r}(p)\neq255\}$ denote the original and output annotated support, with valid background labels included where the dataset defines them. The changed-pixel rate is $100\,|\{p:S_{i,t,r}(p)\neq S_i(p)\}|/(H_iW_i)$, using all image pixels as the denominator. It counts same-location relabeling and support addition or deletion in one edit-footprint measure. We additionally report the changed-pixel footprint divided by $|U_i|$. This support-normalized quantity can exceed $100\%$ when a transformation changes more pixels than the original support area. The full-image rate describes the fraction of the raster changed, while the support-normalized rate describes the scale of that edit relative to sparse input supervision. Neither alone measures semantic or geometric importance.

We also report $R_{\mathrm{class}}$, $R_{\mathrm{offset}}$, and $R_{\mathrm{direction}}$ on that same output. $R_{\mathrm{class}}=|\{p\in U_i\cap V_{i,t,r}:S_{i,t,r}(p)=S_i(p)\}|/|U_i\cap V_{i,t,r}|$ is same-position label retention on common annotated support. It is not GT semantic accuracy. $R_{\mathrm{offset}}$ is the report-defined Offset score on the current output divided by the same image's original-score denominator. $R_{\mathrm{direction}}$ uses fixed original component--GT bindings for non-Direction outputs, while Direction's own target column uses the generator-native Direction score. The S4Pascal side-effect $R_{\mathrm{direction}}$ uses pixel-weighted component aggregation. Manual VOC and S4Cityscapes use their report's fixed-binding estimator. Cityscapes Class $R_{\mathrm{offset}}$ is the fixed-binding geometric control, so class relabeling alone leaves that geometry score unchanged. Ratios are computed per image and then averaged equally over images with a valid denominator. The score ratios have different kernels, are compared with their own no-change reference, and may exceed one. Values above one are retained and are not probabilities or preservation percentages. We report original-position support overlap separately. This measures how much of the input support remains at the same coordinates and can decrease under translation or rotation even when the total support area is unchanged.

Table~\ref{tab:operator-confounding-endpoint} reports the strongest evaluated nominal setting, $r=0.1$, and Figure~\ref{fig:operator-side-effects} shows non-target score ratios over all evaluated levels. Unless stated otherwise, the ranges below are the minimum and maximum across the three dataset rows in Table~\ref{tab:operator-confounding-endpoint} at $r=0.1$. Class changes 2.05--2.90\% of image pixels, while its geometry score ratios and original-position support overlap remain at one. Its same-position class-retention ratio is 0.067--0.093. Offset changes 2.75--4.16\% of pixels and its own score ratio is 0.138--0.304. Its class-retention ratio is 0.853--0.960, while its Direction ratio is 0.964--1.001. Direction changes 1.53--2.20\% of image pixels. Its target score ratio is 0.095--0.106, while class retention is 0.994--0.996 and Offset ratios are 0.908--0.977. Shrink has original-position overlap of 0.100--0.135 and changes 2.03--2.74\% of image pixels. Its class-retention ratio remains near one, while Offset and Direction ratios vary across datasets. For example, the manual-VOC Shrink Direction ratio is 1.067. This is an increase in the reported score ratio for that dataset, not evidence that Shrink generally improves direction. Relative to original scribble support, the mean changed-pixel footprint at $r=0.1$ is 90.7--93.3\% for Class, 106.6--158.8\% for Offset, 67.0--71.7\% for Direction, and 86.5--90.0\% for Shrink across the three datasets. This support-normalized quantity can exceed 100\% because the edit footprint includes changed pixels outside the original support.

\begin{table*}[!t]
\centering\small
\setlength{\tabcolsep}{5pt}

\caption{Cross-property measurements on the $r=0.1$ target-operator outputs. Changed pixels use the full image area as denominator, and changed/support uses the original annotated support area as denominator. The latter can exceed 100\% because edits may affect pixels outside original support. $R_{\mathrm{class}}$ is same-position label retention, $R_{\mathrm{offset}}$ is the report-defined Offset score ratio (except Cityscapes Class, which uses the fixed-binding geometric control), and $R_{\mathrm{direction}}$ is the report-defined fixed-binding ratio for side effects or the generator-native target ratio for Direction outputs. Values are computed per image and averaged equally over images with valid denominators. The final column reports original-position support overlap, $|U\cap V|/|U|$, where $U$ and $V$ denote the input and output labeled support. This measures spatial overlap with the input support, not total output support area; translation can reduce it without deleting support. Ratios above one are retained and are not probabilities. Direction side-effect scores use fixed original component--GT binding with pixel-weighted component aggregation for S4Pascal. These score ratios are not interchangeable physical units.}
\label{tab:operator-confounding-endpoint}
\resizebox{\textwidth}{!}{%
\begin{tabular}{llrrrrrr}
\toprule
Dataset & Target operator & Changed pixels (\%) & Changed/support (\%) & $R_{\mathrm{class}}$ & $R_{\mathrm{offset}}$ & $R_{\mathrm{direction}}$ & Original-position overlap \\
\midrule
VOC2012 (manual) ($n=12{,}031$) & Class & 2.90 & 92.43 & 0.076 & 1.000 & 1.000 & 1.000 \\
VOC2012 (manual) ($n=12{,}031$) & Offset & 4.16 & 132.20 & 0.897 & 0.283 & 1.001 & 0.337 \\
VOC2012 (manual) ($n=12{,}031$) & Direction & 2.20 & 66.97 & 0.995 & 0.908 & 0.095 & 0.654 \\
VOC2012 (manual) ($n=12{,}031$) & Shrink & 2.74 & 87.07 & 0.998 & 0.768 & 1.067 & 0.130 \\
\addlinespace[2pt]
S4Pascal (auto) ($n=12{,}031$) & Class & 2.41 & 93.28 & 0.067 & 1.000 & 1.000 & 1.000 \\
S4Pascal (auto) ($n=12{,}031$) & Offset & 2.75 & 106.59 & 0.960 & 0.304 & 0.995 & 0.465 \\
S4Pascal (auto) ($n=12{,}031$) & Direction & 1.96 & 71.71 & 0.996 & 0.953 & 0.099 & 0.633 \\
S4Pascal (auto) ($n=12{,}031$) & Shrink & 2.25 & 86.53 & 1.000 & 0.969 & 0.960 & 0.135 \\
\addlinespace[2pt]
S4Cityscapes (auto) ($n=2{,}975$) & Class & 2.05 & 90.73 & 0.093 & 1.000 & 1.000 & 1.000 \\
S4Cityscapes (auto) ($n=2{,}975$) & Offset & 3.57 & 158.77 & 0.853 & 0.138 & 0.964 & 0.174 \\
S4Cityscapes (auto) ($n=2{,}975$) & Direction & 1.53 & 71.28 & 0.994 & 0.977 & 0.106 & 0.628 \\
S4Cityscapes (auto) ($n=2{,}975$) & Shrink & 2.03 & 90.00 & 1.000 & 0.787 & 0.963 & 0.100 \\
\bottomrule
\end{tabular}
}
\end{table*}

Taken together, the results show that the operators are not perfectly independent, but the measured off-target score changes are generally limited for Class and Direction and modest for most Offset and Shrink comparisons. We use a descriptive band from 0.95 to 1.05 around one as a practical near-no-change range for score ratios. Across the 27 dataset-by-ratio aggregate cells, all Class-output Offset and Direction ratios and all Direction-output Class ratios fall within this band. For Offset outputs, 27 of 27 Direction ratios and 20 of 27 Class-retention ratios fall within it. For Shrink outputs, 27 of 27 Class-retention ratios and 26 of 27 Direction ratios fall within it. This supports a positive but qualified interpretation that most measured off-target score effects are small under this descriptive criterion. It does not imply that support is unchanged. Offset can alter support location, whereas Shrink targets support extent. The reported overlap measure does not by itself distinguish relocation from total-area reduction. Some score ratios also depart from the band. The band is not a statistical equivalence test, and these operator-specific ratios are not probabilities. Class relabeling leaves the measured geometry and support unchanged. At the strongest setting, Direction outputs retain Class at 0.994--0.996 and have Offset ratios of 0.908--0.977. Offset outputs have Direction ratios of 0.964--1.001 and Class retention of 0.853--0.960. Shrink shows the clearest dataset-dependent geometric side effects while carrying out its intended support reduction. Figure~\ref{fig:operator-side-effects} is the core evidence because it compares non-target properties with their no-change reference across strength levels. Table~\ref{tab:operator-confounding-endpoint} complements the figure with changed-pixel and support measures at the strongest setting. The binding and score-kernel choices, component matching, and dataset-specific annotation structure limit exact cross-dataset comparison. These measurements quantify co-occurring label changes and should not be read as proof that a particular side effect independently causes a change in model performance.

\section{Experimental Protocol}
\subsection{Dataset Structure}
We use manual VOC2012~\citep{everingham2010pascal} scribbles as the primary reference setting and extend
the study with two Scribble4All datasets~\citep{boettcher2024scribblesforall}.
S4Pascal~\citep{boettcher2024scribblesforall} uses automatically generated scribbles on the same PASCAL VOC images
and dense masks, isolating the effect of annotation source while preserving the
image domain. S4Cityscapes~\citep{boettcher2024scribblesforall,cordts2016cityscapes} uses automatically generated scribbles in a distinct
urban-scene domain with different resolution, scene layout, class set, and
scribble density. Together, these settings test whether the observed robustness
patterns persist across manual versus GT-derived scribbles and under broader
dataset and scene-structure shifts.

\subsection{Evaluation Pipelines}
\textbf{TEL} (Tree Energy Loss) treats labeled scribble pixels as sparse anchors and constructs a tree-energy regularizer through affinity filtering and tree-structured propagation~\citep{liang2022tree}.  It is therefore a useful primary pipeline for testing whether a controlled alteration to class identity, position, trajectory, or support changes the supervision propagated beyond the stroke.  

\noindent \textbf{Adapted DTS} (Dynamic Teacher Switching) is a distinct teacher--student pipeline: two exponential moving average (EMA) teachers are assessed by their scribble loss, the selected teacher supplies confidence-filtered pseudo-labels, and hierarchical feature consistency regularizes the student~\citep{nguyen2026sdtnet}.  Only the selected teacher is EMA-updated in an iteration.  We use DTS as an independent pipeline-sensitivity test, not as a strict line-by-line reimplementation or a matched-protocol leaderboard comparison. Its adapted architecture, losses, update order, and inference path are diagrammed in the supplementary material.

\noindent \textbf{PFA} (Prototype-Based Feature Augmentation) provides a third independent pipeline: class prototypes derived from labeled pixels augment the feature supervision available beyond the sparse marks~\citep{chan2024pfa}.  We include its VOC2012--MiT-B1 results as a compact external check on whether the observed operation response persists under a prototype-based learning mechanism.

\subsection{Implementation Details}

\noindent \textbf{TEL.}
We use TEL with DeepLabV3+--ResNet101~\citep{liang2022tree,chen2018deeplabv3plus,he2016resnet}.
VOC2012 and S4Pascal are trained for \(40\mathrm{k}\) iterations with \(512\times512\) crops, batch size \(4\), and SGD (\(\mathrm{lr}=10^{-3}\), momentum \(0.9\), weight decay \(10^{-4}\)), using polynomial decay and a \(2\mathrm{k}\)-iteration warm-up.
S4Cityscapes uses \(512\times1024\) crops, batch size \(2\), \(\mathrm{lr}=0.006\), and \(40\mathrm{k}\) iterations.
We set \(\lambda=0.4\) throughout, with \(\sigma=0.02\) for VOC2012 and S4Pascal and \(\sigma=0.002\) for S4Cityscapes, and use standard PyTorch BatchNorm rather than SyncBN. SegFormer-B4 replaces only the backbone under the same TEL protocol~\citep{xie2021segformer}.

\noindent \textbf{DTS.}
The adapted DTS uses \(512\times512\) crops, batch size \(2\), and SGD (\(\mathrm{lr}=0.00125\), momentum \(0.9\), weight decay \(10^{-4}\)). 
VOC2012 and S4Pascal are trained for \(30\mathrm{k}\) iterations and evaluated every \(2\mathrm{k}\). For ResNet101, the pseudo-label threshold decreases from \(0.90\) to \(0.80\) over \(12\mathrm{k}\) iterations. The pseudo-label and hierarchical feature consistency (HiCo) weights are \(0.0125\) and \(0.05\), with warm-up/ramp schedules of \(2\mathrm{k}/4\mathrm{k}\) and \(4\mathrm{k}/4\mathrm{k}\), respectively, and EMA rates of \(0.999\) and \(0.9999\).
S4Cityscapes uses \(40\mathrm{k}\) iterations, weights \(0.0700/0.10\), and ramps of \(4\mathrm{k}/8\mathrm{k}\) and \(8\mathrm{k}/8\mathrm{k}\), while retaining the same threshold. SegFormer-B4 uses a \(0.85\!\rightarrow\!0.80\) threshold ramp over \(8\mathrm{k}\) iterations.

\noindent \textbf{PFA.}
PFA follows the released MiT-B1 configuration~\citep{chan2024pfa}: \(21\) classes, \(320\times320\) crops, batch size \(12\), and Adam (\(\mathrm{lr}=6\times10^{-5}\), betas \(0.9/0.999\), weight decay \(0.01\)) with linear decay.
It uses ten \(256\)-D prototypes, update rate \(0.99\), prototype activation from epoch \(15\), class threshold \(100\), and global-branch weights \((1,0.01,0.01)\). We change only the training duration, reducing it from \(200\) to \(100\) epochs.

\noindent \textbf{Reproducibility and evaluation.}
The reported runs use single-GPU training and the primary experiments were executed on an NVIDIA A100 GPU (40 GB) with Python 3.10.20, PyTorch 2.0.1, CUDA 11.8, cuDNN 8.7, TorchVision 0.15.2, and MMEngine 0.10.7. The primary TEL--VOC2012--ResNet101 sweep uses seeds \(\{0,1,2,3,4\}\); extension seed counts are listed in Tables~\ref{tab:multiseed-extensions} and~\ref{tab:anchor-multiseed}. TEL checkpoints are saved every \(1\mathrm{k}\) iterations and summarized over the final \(25\%\). Inference uses seven-scale flip test-time augmentation. DTS is evaluated every \(2\mathrm{k}\) iterations using whole-image inference, and the best validation checkpoint is retained. PFA is evaluated once per epoch after prototype warm-up; its global branch starts at epoch \(16\), and the checkpoint with the highest validation mIoU is selected. Validation is used only for checkpoint selection and matched clean/noisy evaluation; it does not provide training targets or modify the generated noise conditions.

\noindent\textbf{Metrics.}
The primary metric is mIoU.
TEL reports the mean over the last quarter of validation checkpoints (tail-$25\%$).
DTS and PFA report the best validation mIoU and use the corresponding best checkpoint.
Let $\mathcal R=\{0.9,0.8,\ldots,0.1\}$ denote the nine non-clean ratios. We summarize the clean-normalized response as
\begin{equation}
\begin{aligned}
C(r)&=\left|\frac{\mathrm{mIoU}(1.0)-\mathrm{mIoU}(r)}
{\mathrm{mIoU}(1.0)}\right|,\\
C_{\mathrm{avg}}&=\frac{1}{|\mathcal R|}\sum_{r\in\mathcal R} C(r),\\
C_{\mathrm{end}}&=C(0.1).
\end{aligned}
\label{eq:sensitivity}
\end{equation}
Larger $C(r)$, $C_{\mathrm{avg}}$, or $C_{\mathrm{end}}$ denotes a larger deviation from that setting's clean reference.
These summaries compare response magnitude within a setting; they do not claim cross-operation severity equivalence.
Repeated-seed evaluation uses a full five-seed primary sweep, then matched anchors on the dataset, backbone, and pipeline axes to test whether the primary ordering recurs beyond the reference setting. 
Their exact conditions and results are reported in Tables~\ref{tab:multiseed-extensions} and~\ref{tab:anchor-multiseed}.

\section{Results}

\subsection{Primary manual-VOC study}
\begin{figure}[t]
\centering
\includegraphics[width=0.98\columnwidth]{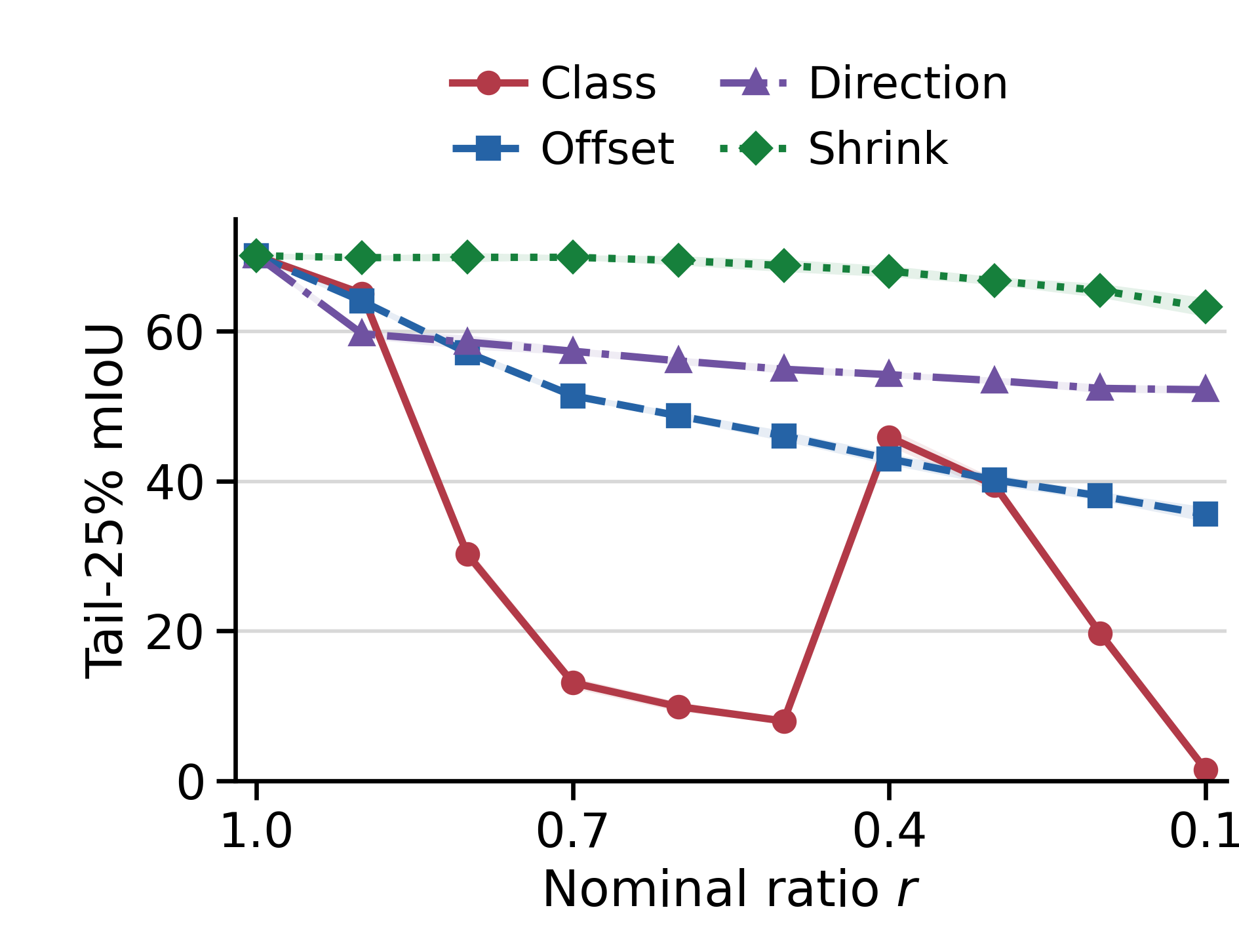}
\caption{Five-seed Manual-VOC TEL--ResNet101 response curves for the four retained operations. Lower \(r\) denotes a stronger within-operator intervention. Shaded bands show $\pm1$ standard deviation across five seeds}
\label{fig:primary}
\end{figure}

Figure~\ref{fig:primary} presents the response curves for the four retained operations in manual-VOC TEL--ResNet101. Random dropout and fragmentation remain secondary single-run controls.

\paragraph{Class corruption.}
Class corruption causes the largest degradation, reaching a five-seed mean of $1.46$ mIoU in its strongest setting, compared with a clean-reference mIoU of $70.08$.
Its curve is strongly non-monotonic: performance first drops sharply, temporarily rebounds at an intermediate corruption level, and then decreases again under stronger corruption.  The rebound is an observed response shape, not evidence of genuine recovery from incorrect semantic supervision.

One possible explanation is that highly unreliable scribbles cause the model to rely relatively more on image features and TEL's structural constraints.  The validation trajectories provide indirect support because the loss becomes less stable under stronger corruption, but they do not establish this mechanism or explain the rebound conclusively.

\paragraph{Spatial offset.}
The spatial offset is the strongest geometric perturbation, reaching a five-seed mean of $35.57$ mIoU at its strongest setting.  Its degradation is consistent with attaching class evidence to unsuitable locations.
The resulting geometric side effects are retained in the reported response because they form part of the complete spatial intervention.

\paragraph{Directional perturbation.}
Directional perturbation produces a weaker response than class corruption and spatial offset, with a five-seed mean of $52.20$ mIoU at its strongest setting.
This result indicates that trajectory changes are harmful, while remaining class and spatial information can still provide useful supervision.

\paragraph{Stroke-length reduction.}
Stroke-length reduction is comparatively mild, with a five-seed mean of $63.24$ mIoU at its strongest setting. Correctly labeled support reduction is less damaging than changing the semantic identity or spatial anchoring of the scribble when some meaningful support remains.

\paragraph{Secondary controls.}
Using the single-run tail-$25\%$ mIoU summaries, we compute the maximum absolute deviation across $r=0.1$--$0.9$ from each control's own clean reference. The deviation is $0.74$ mIoU percentage points for Random Dropout ($70.52$ clean; minimum $69.78$ at $r=0.6$) and $0.86$ points for Fragmentation ($70.31$ clean; minimum $69.44$ at $r=0.4$). These are descriptive single-run results and are separate from the five-seed primary results.
Their weaker initial effects and status as diagnostic operators motivate retaining them as secondary VOC2012 controls rather than repeating them in the cross-setting matrix.

At $r=0.1$, the mean mIoU decreases from the clean reference of $70.08$ to $1.46$, $35.57$, $52.20$, and $63.24$ for Class, Offset, Direction, and Shrink, respectively. These correspond to drops of $68.62$, $34.51$, $17.88$, and $6.84$ mIoU percentage points. The resulting sensitivity ordering is Class, Offset, Direction, and Shrink, from largest to smallest degradation.

\begin{figure*}[!t]
\centering
\includegraphics[width=0.88\textwidth]{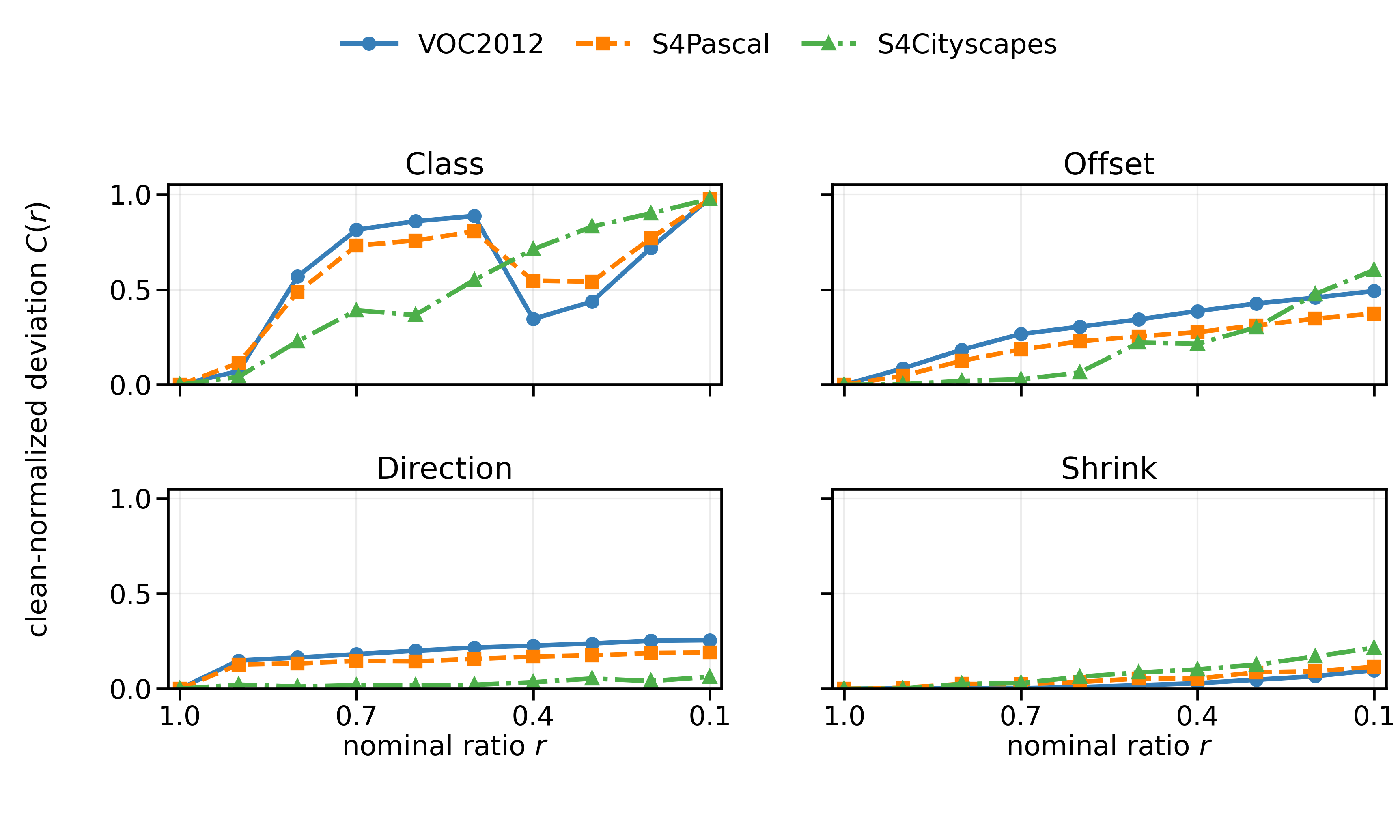}
\caption{TEL--ResNet101 cross-dataset response paths. Each panel shows clean-normalized deviation $C(r)$. VOC2012 shows the five-seed mean path, while S4Pascal and S4Cityscapes show descriptive single-run full paths. Lower \(r\) denotes stronger perturbation within each operator}
\label{fig:cross-dataset-paths}
\end{figure*}

\subsection{Cross-setting robustness}

\paragraph{Cross-dataset and annotation-source robustness.}

Figure~\ref{fig:cross-dataset-paths} compares the clean-normalized response curves across datasets.

We compare TEL with ResNet101 on VOC2012, S4Pascal, and S4Cityscapes.
The curves show clean-normalized response $C(r)$ over the ratio sweep; larger values indicate greater sensitivity.

The associated $C_{\mathrm{avg}}/C_{\mathrm{end}}$ are given in Table~\ref{tab:crosssetting-summary}.

\begin{table*}[!t]
\centering
\scriptsize
\setlength{\tabcolsep}{3.1pt}

\caption{Descriptive TEL response across datasets and backbones. R and S denote ResNet-101 and SegFormer-B4. $C_{\mathrm{avg}}/C_{\mathrm{end}}$ are the average clean-normalized response over $r\in\{0.9,\ldots,0.1\}$ and the response at $r=0.1$. Higher values indicate larger deviation from the clean reference. The table provides numerical summaries for Figure~\ref{fig:cross-dataset-paths}. VOC2012--ResNet101 uses five-seed mean paths. The remaining full-path extensions are descriptive single-run summaries rather than statistical backbone comparisons.}
\label{tab:crosssetting-summary}
\begin{tabular*}{\textwidth}
{@{\extracolsep{\fill}}l cc cc cc@{}}
\toprule
& \multicolumn{2}{c}{VOC2012} & \multicolumn{2}{c}{S4Pascal} & \multicolumn{2}{c}{S4Cityscapes} \\
\cmidrule(lr){2-3}\cmidrule(lr){4-5}\cmidrule(lr){6-7}
Operation & R $C_{\mathrm{avg}} / C_{\mathrm{end}}$ & S $C_{\mathrm{avg}} / C_{\mathrm{end}}$ & R $C_{\mathrm{avg}} / C_{\mathrm{end}}$ & S $C_{\mathrm{avg}} / C_{\mathrm{end}}$ & R $C_{\mathrm{avg}} / C_{\mathrm{end}}$ & S $C_{\mathrm{avg}} / C_{\mathrm{end}}$ \\
\midrule
Class     & 0.631 / 0.979 & 0.632 / 0.967 & 0.636 / 0.976 & 0.643 / 0.979 & 0.555 / 0.976 & 0.597 / 0.980 \\
Offset    & 0.328 / 0.493 & 0.378 / 0.513 & 0.239 / 0.374 & 0.268 / 0.383 & 0.215 / 0.603 & 0.204 / 0.585 \\
Direction & 0.209 / 0.255 & 0.280 / 0.328 & 0.159 / 0.190 & 0.178 / 0.214 & 0.032 / 0.063 & 0.023 / 0.050 \\
Shrink    & 0.031 / 0.098 & 0.034 / 0.041 & 0.055 / 0.116 & 0.034 / 0.056 & 0.091 / 0.215 & 0.042 / 0.109 \\
\bottomrule
\end{tabular*}

\end{table*}

Comparing VOC2012 with S4Pascal isolates annotation-source effects. They share the same PASCAL images and dense ground-truth masks, but use manual and GT-derived scribbles, respectively. Comparing S4Pascal with S4Cityscapes probes broader dataset and scene-structure effects. Both use the automatic Scribble4All pipeline, while differing in domain, resolution, scene layout, class set, and scribble density.

Class corruption is similarly damaging on VOC2012 and S4Pascal. S4Pascal is generally less sensitive to spatial offset and directional perturbation, which may reflect the greater spatial regularity of its GT-derived scribbles. Stroke-length reduction shows a small reversal between the two datasets. A rough length proxy, computed as the published labeled-area fraction divided by the published mean scribble count and an assumed width of approximately $4$ pixels~\citep{boettcher2024scribblesforall}, is about $21\%$ higher for VOC2012 than for S4Pascal. This ratio of aggregate statistics is not the mean of per-image stroke lengths and should be interpreted only as a descriptive estimate. It may partly explain the greater tolerance to stroke-length reduction.

Table~\ref{tab:crosssetting-summary} quantifies these differences. With ResNet101, the $C_{\mathrm{avg}}$ for Class is similar on manual VOC2012 and S4Pascal, at $0.631$ and $0.636$, respectively, while the Direction value decreases from $0.209$ to $0.159$ and then to $0.032$ on S4Cityscapes. For Offset, S4Cityscapes has a lower $C_{\mathrm{avg}}$ of $0.215$ than VOC2012 and S4Pascal, but its $C_{\mathrm{end}}$ reaches $0.603$, compared with $0.493$ and $0.374$. This contrast shows why the average response and the strongest-setting response should be read together. These are descriptive comparisons across settings and do not isolate the cause of the differences.

S4Cityscapes is more tolerant of directional perturbation but more sensitive to spatial offset and stroke-length reduction. This response is compatible with its urban layouts, thin structures, adjacent objects, and different scribble density. No dataset is uniformly more robust: the response depends jointly on scribble source, scene structure, density, and perturbation type.

\paragraph{Cross-backbone robustness.}
We compare TEL with DeepLabV3+--ResNet101 and SegFormer-B4 on VOC2012, S4Pascal, and S4Cityscapes under matched operation settings.
Table~\ref{tab:crosssetting-summary} reports $C_{\mathrm{avg}}/C_{\mathrm{end}}$, where larger values indicate greater sensitivity and smaller values indicate greater robustness. Comparisons are therefore made within the same dataset and operation rather than across different noise types.

Class corruption remains the strongest perturbation for both backbones, and changing the backbone does not remove the semantic-anchor risk. SegFormer-B4 is generally more tolerant of stroke-length reduction on the automatically generated datasets, possibly reflecting its ability to exploit remaining sparse support through multi-scale token representations. ResNet101 is generally more robust to spatial offset and directional perturbation on VOC2012 and S4Pascal, whereas SegFormer-B4 has a small advantage on S4Cityscapes. Backbone robustness is therefore conditional on both the perturbation and the dataset.

\paragraph{Cross-pipeline robustness.}
To examine robustness across learning mechanisms, we compare three scribble-supervision pipelines: affinity-guided TEL, teacher--student DTS, and prototype-augmented PFA. This comparison helps distinguish recurring responses from pipeline-specific ones.

\begin{figure}[t]
\centering
\includegraphics[width=0.98\columnwidth]{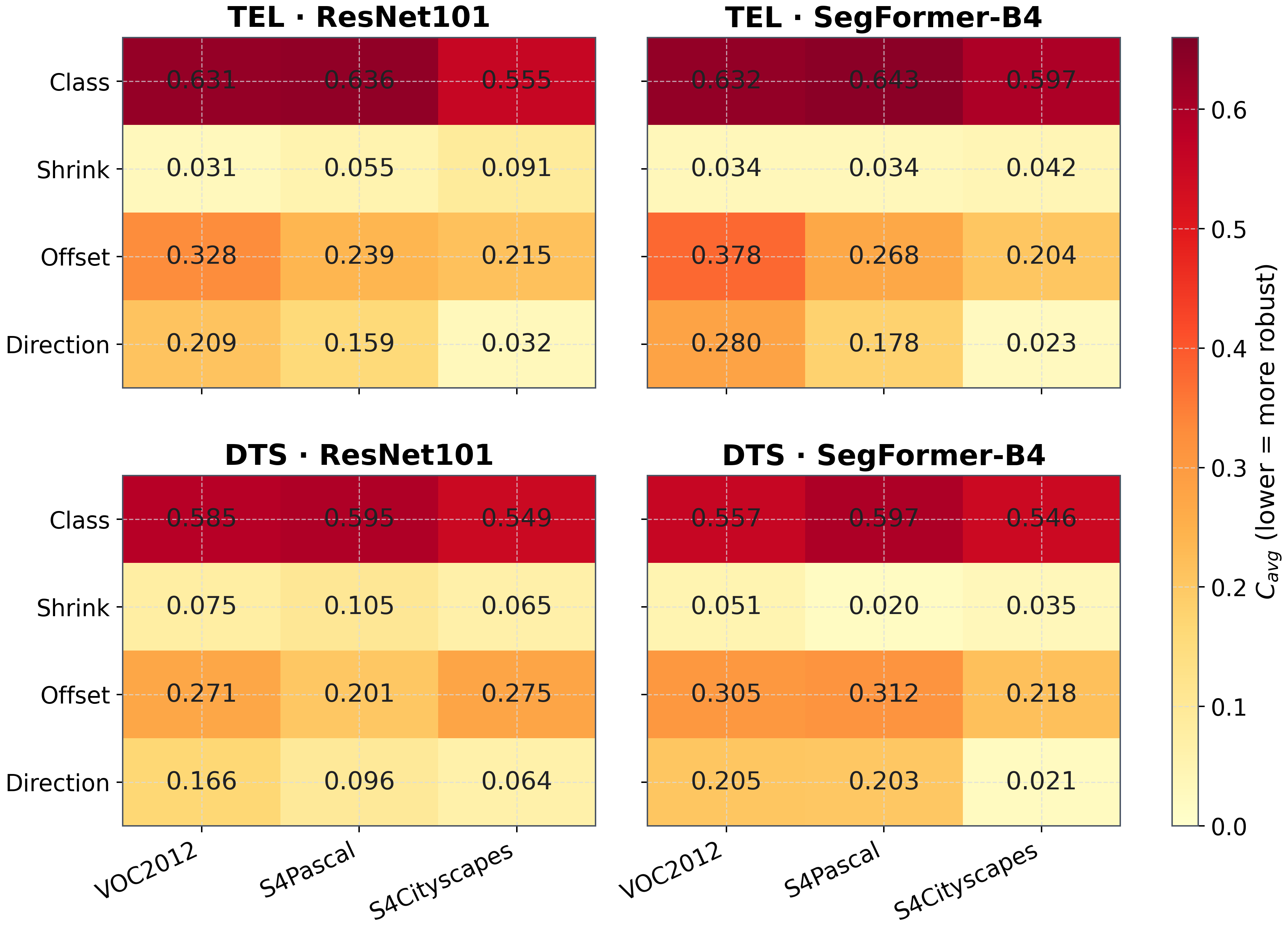}
\caption{Pipeline sensitivity across matched TEL and DTS settings, measured by $C_{\mathrm{avg}}$ over the nine perturbed ratios. TEL--ResNet101 on VOC2012 uses the five-seed mean curve. All other settings, including DTS, use single-run full-ratio sweeps. Darker cells indicate greater clean-normalized response}
\label{fig:pipeline-heatmap}
\end{figure}

Figure~\ref{fig:pipeline-heatmap} summarizes the corresponding clean-normalized sensitivities across the matched TEL and DTS settings. Darker cells indicate larger deviations from the clean reference.
Relative to TEL, DTS is less sensitive to class corruption in several settings and to selected spatial-offset and directional-perturbation conditions. Its response to stroke-length reduction and to some S4Cityscapes conditions is less favorable. Thus, the TEL--DTS contrast is setting-dependent rather than a universal pipeline ranking.

We further evaluate PFA~\citep{chan2024pfa} on VOC2012 with a MiT-B1 backbone and three seeds at the clean and \(r=0.1\) anchor points (Table~\ref{tab:anchor-multiseed}). At the evaluated endpoints, class corruption and spatial offset produce the two largest mIoU drops across the three pipelines, while the ordering of directional perturbation and stroke-length reduction depends on the pipeline. This anchor-based result provides an additional cross-pipeline check.

\begin{table*}[t]
\centering
\scriptsize
\setlength{\tabcolsep}{3.0pt}
\renewcommand{\arraystretch}{0.94}

\caption{Repeated-seed results for the three settings with four reported anchor ratios (percentages; mean $\pm$ sample SD). Pipeline, backbone, and dataset are separated explicitly. TEL entries use the tail-$25\%$ validation average, whereas DTS entries use the best validation checkpoint under its logged protocol. Repeated $r{=}1.0$ values are the shared clean reference within each setting.}
\label{tab:multiseed-extensions}
\begin{tabular*}{\textwidth}{@{\extracolsep{\fill}}p{0.08\textwidth}p{0.11\textwidth}p{0.11\textwidth}p{0.10\textwidth}c >{\centering\arraybackslash}p{0.12\textwidth} >{\centering\arraybackslash}p{0.12\textwidth} >{\centering\arraybackslash}p{0.12\textwidth} >{\centering\arraybackslash}p{0.12\textwidth}@{}}
\toprule
\multirow{2}{*}{Pipeline} & \multirow{2}{*}{Backbone} & \multirow{2}{*}{Dataset} & \multirow{2}{*}{Noise} & \multirow{2}{*}{$n$} & \multicolumn{4}{c}{Validation mIoU (mean $\pm$ sample SD)}\\
\cmidrule(l){6-9}
 & & & & & $r{=}1.0$ & $r{=}0.9$ & $r{=}0.5$ & $r{=}0.1$\\
\midrule
\multirow{4}{*}{TEL} & \multirow{4}{*}{ResNet101} & \multirow{4}{*}{VOC2012} & Class & 5 & $70.08\pm0.42$ & $65.04\pm0.61$ & $7.96\pm0.25$ & $1.46\pm0.12$\\
 & & & Offset & 5 & $70.08\pm0.42$ & $64.03\pm0.39$ & $46.03\pm0.68$ & $35.57\pm0.98$\\
 & & & Direction & 5 & $70.08\pm0.42$ & $59.66\pm0.59$ & $54.94\pm0.52$ & $52.20\pm0.44$\\
 & & & Shrink & 5 & $70.08\pm0.42$ & $69.83\pm0.27$ & $68.77\pm0.82$ & $63.24\pm1.25$\\
\midrule
\multirow{4}{*}{TEL} & \multirow{4}{*}{ResNet101} & \multirow{4}{*}{S4Pascal} & Class & 4 & $75.76\pm0.17$ & $69.43\pm0.19$ & $13.07\pm0.54$ & $1.46\pm0.10$\\
 & & & Offset & 4 & $75.76\pm0.17$ & $74.28\pm0.22$ & $60.10\pm0.45$ & $52.06\pm0.34$\\
 & & & Direction & 4 & $75.76\pm0.17$ & $70.41\pm0.07$ & $67.01\pm0.27$ & $64.29\pm0.25$\\
 & & & Shrink & 4 & $75.76\pm0.17$ & $75.08\pm0.24$ & $71.39\pm0.26$ & $68.79\pm0.49$\\
\midrule
\multirow{4}{*}{DTS} & \multirow{4}{*}{ResNet101} & \multirow{4}{*}{VOC2012} & Class & 4 & $55.73\pm0.92$ & $43.41\pm0.71$ & $9.95\pm0.61$ & $5.91\pm0.22$\\
 & & & Offset & 4 & $55.73\pm0.92$ & $53.34\pm0.57$ & $39.57\pm0.43$ & $32.70\pm0.53$\\
 & & & Direction & 4 & $55.73\pm0.92$ & $49.78\pm0.26$ & $46.33\pm0.49$ & $44.29\pm0.59$\\
 & & & Shrink & 4 & $55.73\pm0.92$ & $55.67\pm0.16$ & $52.53\pm0.56$ & $41.22\pm0.52$\\
\bottomrule
\end{tabular*}

\end{table*}

\begin{table*}[!t]
\centering
\scriptsize
\setlength{\tabcolsep}{2.1pt}
\renewcommand{\arraystretch}{0.94}
\caption{Repeated-seed anchor-only results for the two settings evaluated at clean and strong noise (percentages; mean $\pm$ sample SD). TEL entries use the tail-$25\%$ validation average, whereas PFA entries use the best validation checkpoint. Only clean and endpoint conditions were evaluated; intermediate response shapes are therefore not assessed.}
\label{tab:anchor-multiseed}
\begin{tabular}{@{}lll l ccc@{}}
\toprule
\multirow{2}{*}{Pipeline} & \multirow{2}{*}{Backbone} & \multirow{2}{*}{Dataset} & \multirow{2}{*}{Noise} & \multirow{2}{*}{$n$} & \multicolumn{2}{c}{Validation mIoU (mean $\pm$ sample SD)}\\
\cmidrule(l){6-7}
 & & & & & $r{=}1.0$ & $r{=}0.1$\\
\midrule
\multirow{4}{*}{PFA} & \multirow{4}{*}{MiT-B1} & \multirow{4}{*}{VOC2012} & Class & 3 & $78.32\pm0.10$ & $4.46\pm0.06$\\
 & & & Offset & 3 & $78.32\pm0.10$ & $39.94\pm0.05$\\
 & & & Direction & 3 & $78.32\pm0.10$ & $58.83\pm0.03$\\
 & & & Shrink & 3 & $78.32\pm0.10$ & $73.25\pm0.37$\\
\midrule
\multirow{4}{*}{TEL} & \multirow{4}{*}{SegFormer-B4} & \multirow{4}{*}{VOC2012} & Class & 3 & $73.75\pm0.12$ & $3.87\pm0.12$\\
 & & & Offset & 3 & $73.75\pm0.12$ & $38.90\pm0.17$\\
 & & & Direction & 3 & $73.75\pm0.12$ & $56.10\pm0.50$\\
 & & & Shrink & 3 & $73.75\pm0.12$ & $65.95\pm1.18$\\
\bottomrule
\end{tabular}
\end{table*}

\FloatBarrier

\subsection{Stability across settings}
Repeated-seed experiments cover the main manual-VOC study and targeted extensions along the dataset, backbone, and pipeline axes (Tables~\ref{tab:multiseed-extensions} and~\ref{tab:anchor-multiseed}).

The primary TEL--VOC2012--ResNet101 study uses five independent seeds over the full sweep $r\in\{1.0,0.9,\ldots,0.1\}$, testing whether the response paths---including the non-monotonic class curve---are stable across training randomness.

Additional checks use four seeds for TEL--S4Pascal--ResNet101 at $r\in\{0.9,0.5,0.1\}$ and DTS--VOC2012--ResNet101 at $r\in\{1.0,0.9,0.5,0.1\}$. Three-seed clean/$r=0.1$ comparisons cover TEL--VOC2012--SegFormer-B4 and PFA--VOC2012--MiT-B1. These checks test recurrence of the main pattern. Complete seed records are provided in the supplementary material.

Across settings, Class and Offset have the largest endpoint drops, while Direction and Shrink vary by setting.

\section{Conclusion}
We present an audit-informed benchmark of six scribble perturbations. Semantic corruption causes the largest loss, followed by spatial offset. Direction and Shrink have milder, setting-dependent effects. Robustness depends on sparse-label content and placement.

\paragraph{Limitations and future work.}
Our GT-guided operators enable controlled diagnosis of specific annotation properties. Connecting them to the frequency and causes of human errors is an important next step. The complete multi-seed TEL--VOC study and targeted extensions provide a basis for broader validation. Future work will combine human-annotation studies with these diagnostics to develop adaptive, corruption-aware training across datasets, architectures, and annotation sources.


\bibliography{sn-bibliography}

\end{document}